\documentclass[10pt,twocolumn,letterpaper]{article}

\usepackage[pagenumbers]{cvpr} 

\usepackage{graphicx}
\usepackage{dsfont}
\usepackage{amsmath}
\usepackage{amssymb}
\usepackage{booktabs}

\usepackage{algorithm}
\usepackage{listings}

\usepackage{threeparttable}
\usepackage{multirow}
\usepackage{multicol}
\usepackage{color,colortbl}

\definecolor{Gray}{gray}{0.85}
\newcolumntype{a}{>{\columncolor{Gray}}c}
\definecolor{blush}{rgb}{0.87, 0.36, 0.51}

\usepackage[pagebackref,breaklinks,colorlinks]{hyperref}

\usepackage[capitalize]{cleveref}
\crefname{section}{Sec.}{Secs.}
\Crefname{section}{Section}{Sections}
\Crefname{table}{Table}{Tables}
\crefname{table}{Tab.}{Tabs.}

\def\cvprPaperID{9398} 
\def\confName{CVPR}
\def\confYear{2023}

\begin{document}

\title{Modality-Agnostic Debiasing for Single Domain Generalization}

\author{Sanqing Qu$^{1}$\thanks{This work was performed at JD.com}\ , Yingwei Pan$^{2}$, Guang Chen$^{1}$\thanks{Corresponding author}\ , Ting Yao$^{2}$, Changjun Jiang$^{1}$, Tao Mei$^{2}$\\
{\small $^{1}$Tongji University, $^{2}$JD.com}\\
{\tt \scriptsize \{2011444, guangchen, cjjiang\}@tongji.edu.cn, \{panyw.ustc, tingyao.ustc\}@gmail.com, tmei@live.com} 
}
\maketitle

\begin{abstract}
Deep neural networks (DNNs) usually fail to generalize well to outside of distribution (OOD) data, especially in the extreme case of single domain generalization (single-DG) that transfers DNNs from single domain to multiple unseen domains. Existing single-DG techniques commonly devise various data-augmentation algorithms, and remould the multi-source domain generalization methodology to learn domain-generalized (semantic) features. Nevertheless, these methods are typically modality-specific, thereby being only applicable to one single modality (e.g., image).  In contrast, we target a versatile Modality-Agnostic Debiasing (MAD) framework for single-DG, that enables generalization for different modalities. Technically, MAD introduces a novel two-branch classifier: a biased-branch encourages the classifier to identify the domain-specific (superficial) features, and a general-branch captures domain-generalized features based on the knowledge from biased-branch. Our MAD is appealing in view that it is pluggable to most single-DG models. We validate the superiority of our MAD in a variety of single-DG scenarios with different modalities, including recognition on 1D texts, 2D images, 3D point clouds, and semantic segmentation on 2D images. More remarkably, for recognition on 3D point clouds and semantic segmentation on 2D images, MAD improves DSU by 2.82\% and 1.5\% in accuracy and mIOU.

\end{abstract}

\section{Introduction}
\label{sec:intro}
\label{sec:intro}
\par Deep neural networks (DNNs) have achieved remarkable success in various tasks under the assumption that training and testing domains are independent and sampled from identical or sufficiently similar distribution~\cite{vapnik1991principles, ben2010theory}. However, this assumption often does not hold in most real-world scenarios. When deploying DNNs to unseen or out-of-distribution (OOD) testing domains, inevitable performance degeneration is commonly observed.
The difficulty mainly originates from that the backbone of DNNs extracts more domain-specific (superficial) features together with domain-generalized (semantic) features. Therefore, the classifier is prone to paying much attention to those domain-specific features, and learning unintended decision rule~\cite{robust_representation}. To mitigate this issue, several appealing solutions have been developed, including \emph{Domain Adaptation (DA)}~\cite{tpn_da, mmd, dann, BMD, GLC} and \emph{Domain Generalization (DG)}~\cite{dg_survey, congan_dg, mixstyle, ddg}.  Despite showing encouraging performances on OOD data, their real-world applications are still limited due to the requirement to have the data from other domain (i.e., the unseen target domain or multiple source domains with different distributions). In this work, we focus on an extreme case in domain generalization: \emph{single domain generalization (single-DG)}, in which DNNs are trained with single source domain data and then required to generalize well to multiple unseen target domains.
\begin{figure}[t]
    \centering
    \vspace{-0.00in}
    \includegraphics[width=0.47\textwidth]{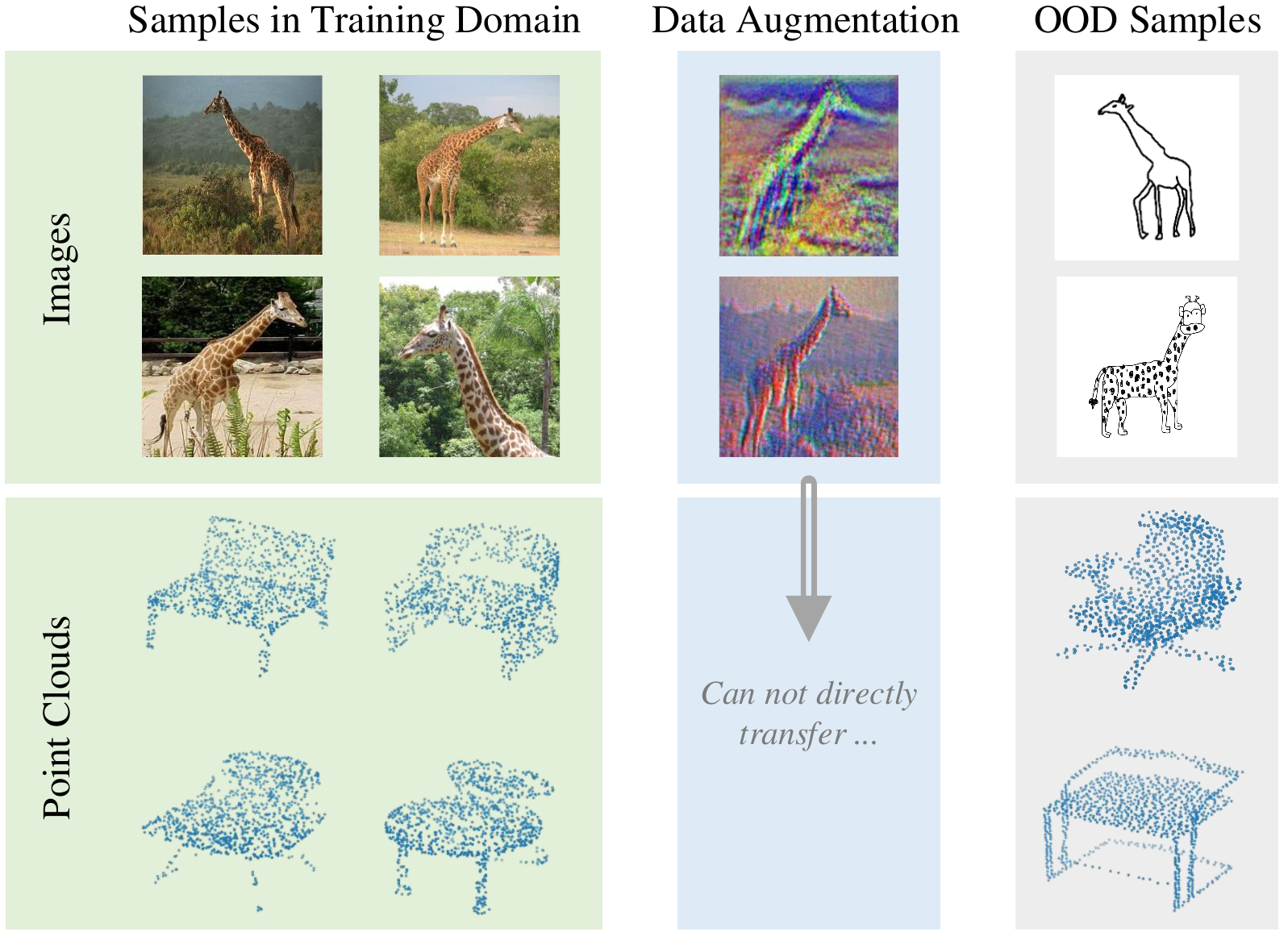}
    \vspace{-0.12in}
    \caption{Most existing single-DG techniques devise various data augmentation algorithms to introduce various image textures and styles, pursuing the learning of domain-generalized features. However, these approaches are modality-specific, and only applicable to single modality (e.g., image). Hence it is difficult to directly employ such single-DG approach for 3D point clouds, since the domain shifts in 3D point clouds only reflect the geometric differences rather than texture and style differences.}
    \vspace{-0.25in}
    \label{fig:compare_img_point}
\end{figure}

\par Previous researches~\cite{cnn_texture_bias, high_freq} demonstrate that the specific local textures and image styles tailored to each domain are two main causes, resulting in domain-specific features for images. To alleviate this, recent works~\cite{L2A, L2D, DSU, mada_pami} design a variety of data-augmentation algorithms to introduce diversified textures and image styles. The DG methodologies are then remolded with these data-augmentation algorithms to facilitate the learning of domain-generalized features. Nevertheless, such solution for single-DG is typically modality-specific and only applicable to the single modality inputs of images. When coming a new modality (e.g. 3D point clouds), it is difficult to directly apply these techniques to tackle single-DG problem. This is due to the fact that the domain shift in 3D point clouds is interpreted as the differences of 3D structural information among multiple domains, instead of the texture and style differences in 2D images~\cite{pointdan, point_mixup}. Figure~\ref{fig:compare_img_point} conceptually illustrates the issue, which has been seldom explored in the literature.

\par In this paper, we propose to address this limitation from the standpoint of directly strengthening the capacity of classifier to identify domain-specific features, and meanwhile emphasize the learning of domain-generalized features. Such way completely eliminates the need of modality-specific data augmentations, thereby leading to a versatile modality-agnostic paradigm for single-DG. Technically, to materialize this idea, we design a novel Modality-Agnostic Debiasing (MAD) framework, that facilitates single domain generalization under a wide variety of modalities. In particular, MAD integrates the basic backbone for feature extraction with a new two-branch classifier structure. One branch is the biased-branch that identifies those superficial and domain-specific features with a multi-head cooperated classifier. The other branch is the general-branch that learns to capture the domain-generalized representations on the basis of the knowledge derived from the biased-branch. It is also appealing in view that our MAD can be seamlessly incorporated into most existing single-DG models with data-augmentation, thereby further boosting single domain generalization.

\par We analyze and evaluate our MAD under a variety of single-DG scenarios with different modalities, ranging from recognition on 2D images, 3D point clouds, 1D texts, to semantic segmentation on 2D images. Extensive experiments demonstrate the superior advantages of MAD when being plugged into a series of existing single-DG techniques with data-augmentation (e.g., Mixstyle~\cite{mixstyle} and DSU~\cite{DSU}). More remarkably, for recognition on point cloud benchmark, MAD significantly improves DSU in the accuracy from 33.63\% to 36.45\%. For semantic segmentation on image benchmark, MAD advances DSU with mIoU improvement from 42.3\% to 43.8\%.

\begin{figure*}
    \centering
    \vspace{-0.00in}
    \includegraphics[width=0.99\textwidth]{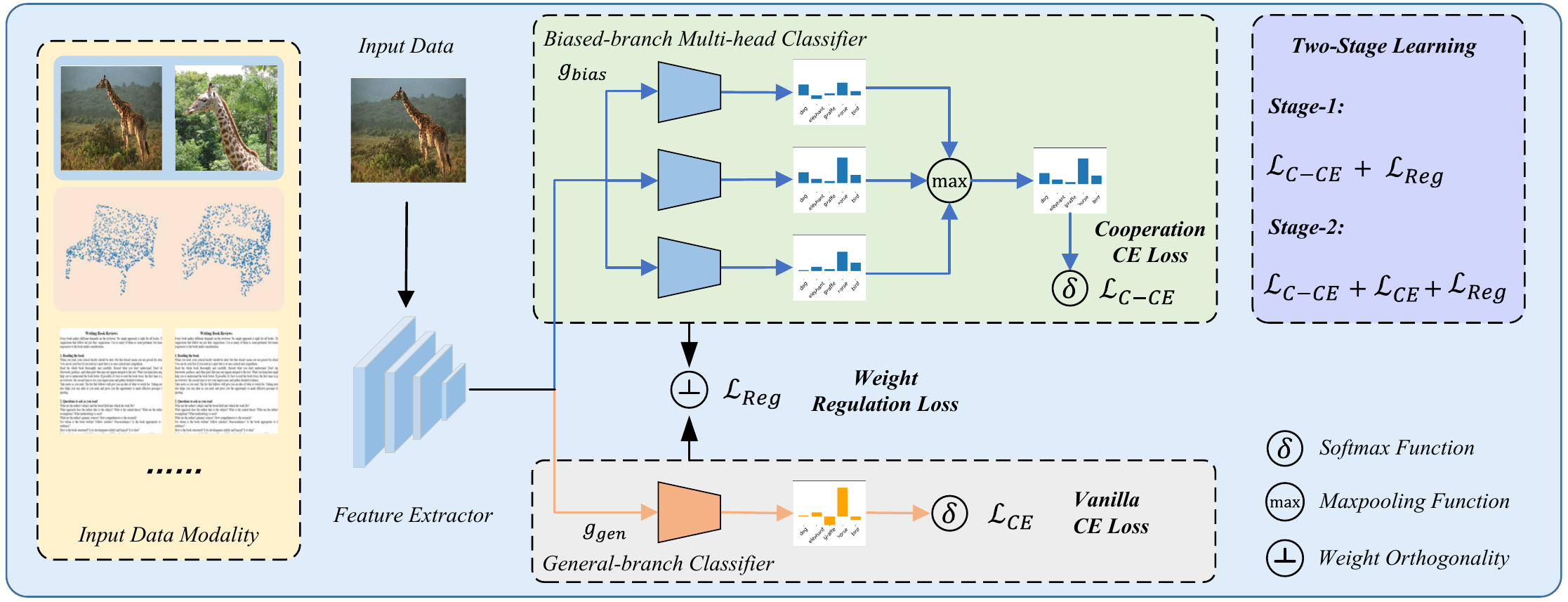}
    \vspace{-0.10in}
    \caption{An overview of our Modality-Agnostic Debiasing (MAD) framework. The main challenge for DNNs to realize single domain generalization (single-DG) is that the classifier tends to over-emphasize those domain-specific features, yielding unintended decision rules~\cite{robust_representation}.
    To address this challenge, we propose to strengthen the capability of classifier of identifying domain-specific features, and meanwhile emphasising the learning of domain-generalized features. Technically, the MAD framework integrates the basic backbone for feature extraction with a new two-branch classifier, i.e., the biased branch and the general branch. We implement MAD in a two-stage learning mechanism. In the first stage, the biased-branch is utilized to identify those domain-specific features with a multi-head cooperated classifier. In the second stage, the general-branch classifier is encouraged to capture those domain-generalized features on the basis of the knowledge from the biased-branch, i.e., with the guidance from $\mathcal{L}_{reg}$. Our framework is modality-agnostic and can be applied to various modalities such as images, point clouds, and texts.}
    \vspace{-0.20in}
    \label{fig:lab_framework}
\end{figure*}

\section{Related Work}

\subsection{Domain Adaptation}
\par Over the last decade, many efforts have been devoted to domain adaptation (DA) to address the OOD issue~\cite{da_survey, mmd, dann}. DA methods are developed to utilize the labeled source domain and the unlabeled out-of-distributed target domain in a transductive learning manner. Existing DA approaches can be briefly grouped into two paradigms, i.e., moment matching~\cite{mmd, wasserstein, dacs_da_seg} and adversarial alignment~\cite{dann, cycada, da_object_1}. DA methods have achieved significant progress in many applications, e.g., object recognition~\cite{BMD, GLC, yao2015semi, pan_osda}, semantic segmentation~\cite{cycada, dacs_da_seg}, and object detection~\cite{da_object_1, da_object_2, MTOR}. Nevertheless, the requirement of both source and target domain data during training significantly limits their practical deployment. Besides, in DA manner, DNNs are typically coupled with source and target domains, affecting their capacity to generalize to other domains. In this work, we focus on a more challenging scenario where DNNs are required to generalize well to multiple unseen domains.

\subsection{Domain Generalization}
\par Different from DA, domain generalization (DG) expects to learn generalized DNNs with the assistance of multiple source domains~\cite{dg_survey, congan_dg, mixstyle}, without the access to target domain. Currently, DG methods can mainly be categorized into three dimensions, including domain alignment, data augmentation/generation, and ensemble learning. Most existing DG methods~\cite{muandet2013domain, ghifary2016scatter, hu2020domain} belong to the category of domain alignment. Their motivation is straightforward: features that are invariant to the source domain shifts should also be generalized to any unseen target domain shift. Data generation is another popular technique for DG~\cite{volpi2019addressing, shi2020towards, xu2020robust}. The goal is to generate diverse and rich data to boost the generalization ability of DNNs. Existing methods typically remould the Variational Auto-encoder (VAE)~\cite{VAE}, and the Generative Adversarial Networks (GAN)~\cite{GAN} to execute diversified data generation. Ensemble learning~\cite{ensemble_learning} commonly learns multiple copies of the same model with different initialization and then utilizes their ensemble for final prediction. As the variant of ensemble learning, weight averaging~\cite{cha2021swad}, domain-specific neural networks~\cite{zhou2021domain}, and domain-specific batch normalization~\cite{seo2020learning} have recently achieved promising results. Nevertheless, it is non-trivial to directly apply these DG techniques for single domain generalization.

\subsection{Single Domain Generalization}
\par Single domain generalization (single-DG) is an extreme case of domain generalization, where DNNs are trained with only one source domain data and required to perform well to multiple unseen target domains. It is more challenging than DA and DG, yet indeed more realistic in practical applications. To address this challenging problem, several methods~\cite{L2D, DSU, mada_pami} have designed various data augmentation algorithms to enhance the diversity and informativeness of training data. In~\cite{L2D}, the authors propose a style-complement module to synthesize images from diverse distributions. In~\cite{DSU}, synthesized feature statistics are introduced to model the uncertainty of domain shifts during training. To regulate single-DG training, \cite{acvc} applies a variety of visual corruptions as augmentation and designs a new attention consistency loss. A novel image meta-convolution network is developed in~\cite{meta_conv} for capturing more domain-generalized features. Nevertheless, most methods are modality-specific and only applicable to image inputs. When we encounter a new data modality, they are commonly not available to deploy. The reason behind is that for different data modalities, domain shifts tend to be different. For example, the differences in 3D geometry structure among multiple domains are the origin of domain shifts for point clouds, instead of style and texture differences in 2D images. Our work delves into this limitation and targets for proposing a general and versatile framework for single-DG that is agnostic to data modality.

\section{Methodology}
\subsection{Preliminary}
\par We consider an extreme case in generalization: single domain generalization (single-DG), where the goal is to train DNNs with single source domain $\mathcal{D}_{S}$ that perform well to multiple unseen target domains: $\{ \mathcal{D}_{T}^{1}, \mathcal{D}_{T}^{2},\dots, \mathcal{D}_{T}^{Z} \}$. In particular, we consider the $K$-way classification. We denote $\mathcal{D}_{S} = \{ (x_i, y_i)\}^{n}_{i=1}$, where $x\in \mathcal{X} \subset \mathbb{R}^{X}, y\in \mathcal{Y} \subset \mathbb{R}^{K}$. The whole DNN architecture is represented as $F = g \circ f$, where $f: \mathbb{R}^{X} \rightarrow \mathbb{R}^{D}$ denotes the feature extractor and $g: \mathbb{R}^{D} \rightarrow \mathbb{R}^K$ is the classifier. This setting is guaranteed under a general assumption in domain generalization: There are domain-generalized features $e_g$ in the domain $\mathcal{D}_S$ whose correlation with label is consistent across domains, and domain-specific features $e_s$ whose correlation with label varies across domains. Classifiers that rely on domain-generalized features $e_g$ perform much better on new unseen domains than those that depend on domain-specific features $e_s$.  In this setting, directly applying the vanilla empirical risk minimization (ERM)~\cite{vapnik1991principles} on $\mathcal{D}_S$ commonly results in a sub-optimal model that does not generalize well to unseen domains. The main reason originates from that the feature extractor $f$ often extracts more domain-specific features $e_s$ together with domain-generalized features $e_g$~\cite{robust_representation}. DNNs trained with SGD often count on the simplest features~\cite{shah2020pitfalls}, which leads to a tendency for the classifier $g$ to overemphasize $e_s$ and pay less attention to $e_g$, resulting in unintended decision rules.

\par Prior methods~\cite{L2D, DSU, mada_pami, acvc} have designed various data-augmentation algorithms to encourage the feature extractor $g$ to learn more domain-generalized features $e_g$ and suppress those domain-specific features $e_s$. However, these algorithms are typically modality-specific, and largely limited to images. Instead, we propose to mitigate this limitation by directly strengthening the capacity of classifier for identifying domain-specific features, and meanwhile emphasising the learning of domain-generalized features. That completely eliminates the requirement of modality-specific data augmentations, pursuing a versatile and modality-agnostic paradigm for single-DG. Technically, we present a novel modality-agnostic debiasing (MAD) framework. MAD integrates the basic backbone for feature extraction with a new two-branch classifier structure. One branch is the biased-branch that identifies those domain-specific features $e_s$ with a multi-head cooperated classifier. The other is that learns to capture the domain-generalized features $e_g$ with the knowledge derived from the biased-branch. Figure~\ref{fig:lab_framework} illustrates the detailed architecture of our MAD.

\subsection{Identifying Domain-specific Features}
\par There have been some efforts~\cite{low_rank_decompose, decompose_1, decompose_2, decompose_3} in domain generalization to realize domain-specific features $e_s$ and domain-generalized features $e_g$ separation.
Nevertheless, most of them require multiple training domains and pre-defined domain labels, making them inapplicable for single-DG. Moreover,~\cite{low_rank_decompose} has pointed out that given a trained classifier, it is non-trivial to uniquely decompose the classifier weight into domain-specific and domain-generalized terms, especially with only one source domain data.

\par To alleviate these issues, we propose a simple yet effective domain-specific feature identification strategy. Our motivation is straightforward: since the vanilla classifier trained with SGD will inadvertently focus more on those domain-specific features, the weights of the trained classifier can be considered as an indicator of those features.

\par Nevertheless, a single vanilla classifier is typically not effective to locate all domain-specific features. The reason is that there commonly exist multiple factors that contribute to domain-specific features. Taking the identification of ``elephants" and ``cats" as an example, the hypotheses ``elephants tend to be found in grasslands", and ``elephants tend to have wrinkled skin" are both beneficial for classification. When we deploy classifiers to the real world, these hypotheses are domain-specific and superficial, and might result in severe performance degradation. For images, there are several factors typically correlated to domain-specific features, such as the background contexts~\cite{background_bias}, the texture of the objects~\cite{cnn_texture_bias}, and high-frequency patterns that are almost invisible to the human eye~\cite{high_freq}. That motivates us to design a biased-branch that identifies more domain-specific features with a multi-head cooperated classifier $g_{bias}: \mathbb{R}^{D} \rightarrow \mathbb{R}^{K\times M}$.  Specifically, we apply the cooperation cross-entropy loss to learn this branch as:
\begin{align}
    & \mathcal{L}_{C-CE} = \mathbb{E}_{x, y} \sum_{k=1}^{K} -\mathds{1}_{[k=y]} \log \frac{\exp(\max(v_k(x)))}{\sum_{j=1}^{K} \exp(\max(v_j(x)))},
    \label{equ:erm_bias}
\end{align}
where $v_k(x)= g_{bias}(f(x))[k, :] \in \mathbb{R}^{M}$ denotes the logits of multi-head classifier for the $k$-th category of sample $x$, and $M$ is the number of classification heads. Note that we do not enforce all heads of the biased-branch classifier to correctly predict each sample. Instead, we only need one of them to accurately identify it. That is, all heads are encouraged to cooperate with each other for classification. The spirit behind is that domain-specific features do not represent the truly domain-generalized semantics. Thereby, for a particular-type domain-specific features, they are not necessarily present in all samples. Since the \emph{max} function is not differentiable in Eq.~(\ref{equ:erm_bias}), we approximate this function with the \emph{log-sum-exp} during our implementation.

\par In general, the sweet spot for $M$ is set within the range from 1 to $D//K - 1$. Its value depends on the dimension and factors introduced domain-specific features. In our implementation, we perform cross validation to choose a good value for $M$, but it is worth noting that the performance is relatively stable with respect to this choice (see more discussions in Sec.~\ref{sec:exp_ablation_hyper_param}).

\subsection{Learning to Debias}
\par Based on the proposed biased-branch, we have an indicator to those domain-specific features. A follow-up question is how to suppress those domain-specific features in favor of focusing more on those desired domain-generalized features. Here, we introduce another general-branch classifier $g_{gen} : \mathbb{R}^{D} \rightarrow \mathbb{R}^{K}$ to capture those domain-generalized features. Let $W_{bias} \in \mathbb{R}^{K\times M\times D}$ and $W_{gen} \in \mathbb{R}^{K\times D}$ be weights of the multi-head biased classifier and the domain-general classifier, respectively. An intuitive solution is to enforce orthogonality between $W_{bias}$ and $W_{gen}$ in Eq.~(\ref{equ:reg_loss}) during learning the classifier $g_{gen}$ in Eq.~(\ref{equ:erm_gen}):
\begin{align}
    & \mathcal{L}_{CE} = \mathbb{E}_{x, y} \sum_{k=1}^{K} -\mathds{1}_{[k=y]} \log \frac{\exp(u_k(x))}{\sum_{j=1}^{K} \exp(u_j(x))},
    \label{equ:erm_gen}
\end{align}
\begin{align}
    &\mathcal{L}_{Reg} = \frac{1}{K} \sum_{k=1}^{K} \left\| W_{bias}[k,:]\ W_{gen}[k,:]^T \right\|_{F}^{2}
    \label{equ:reg_loss}.
\end{align}
\noindent Here $u_k(x)= g_{gen}(f(x))[k] \in \mathbb{R}$ represents the logit of classifier $g_{gen}$ for the $k$-th category of input sample $x$. However, if we optimize the whole network (including the feature extractor $f$, biased-branch classifier $g_{bias}$, and general-branch classifier $g_{gen}$) simultaneously at the beginning, there is no guarantee that the classifier $g_{gen}$ will pay more attention to those domain-general features. To address this issue, we introduce a two-stage learning mechanism to enable the interaction between the two branches. Technically, in the first stage, we only introduce Eq.~(\ref{equ:erm_bias}) and Eq.~(\ref{equ:reg_loss}) to optimize the network, encouraging the biased-branch classifier to learn those domain-specific features and expecting the weight of general-branch classifier $W_{gen}$ to evade the territory of domain-specific features. Then, in the second stage, we apply Eq.~(\ref{equ:erm_bias}), Eq.~(\ref{equ:reg_loss}) and Eq.~(\ref{equ:erm_gen}) together to optimize the entire network. Accordingly, the overall optimization objective is:
\begin{align}
    \min_{f, g_{bias}, g_{gen}} \mathcal{L}_{C-CE} + \mathcal{L}_{Reg} + \mathds{1}_{[pro \ge T]}\cdot \mathcal{L}_{CE},
\end{align}
where all loss terms are equally weighted, $pro$ denotes the overall training progress, $T$ is a hyper-parameter that determines when to trigger the second stage learning. In general, the choice of $T$ depends on the training dataset size and task difficulty. In our implementation, for recognition task, we typically set $T = 3$ epochs (50 epochs in total). As for semantic segmentation, we set $T = 6\%$ of the iterations in total. Algorithm.~\ref{alg:code} presents the Pseudo-code of our MAD.
\begin{algorithm}[t]
\caption{MAD Pseudocode, PyTorch-like}
\label{alg:code}
\definecolor{codeblue}{rgb}{0.25,0.5,0.5}
\definecolor{codekw}{rgb}{0.85, 0.18, 0.50}
\lstset{
  backgroundcolor=\color{white},
  basicstyle=\fontsize{7.5pt}{7.5pt}\ttfamily\selectfont,
  columns=fullflexible,
  breaklines=true,
  captionpos=b,
  commentstyle=\fontsize{7.5pt}{7.5pt}\color{codeblue},
  keywordstyle=\fontsize{7.5pt}{7.5pt}\color{codekw},
}
\begin{lstlisting}[language=python]
# f: feature extractor
# cls_g1: the biased-branch multi-head classifier
# cls_g2: the general-branch classifier
# pro, T: training progress, second stage thresh

# During inference, we only utilize f, and cls_g2

for x, y in loader:  # load a minibatch x, y 
    z = f(x) # (N, D)
    c1 = cls_g1(z) # (N, K, M)
    c1 = logsumexp(c1, dim=-1) # (N, K)
    l_bias = CrossEntropyLoss(c1, y)
    
    c2 = cls_g2(z) # (N, K)
    l_gen = CrossEntropyLoss(c2, y)
    
    g1_w = cls_g1.weight # (K, M, D)
    g2_w = cls_g2.weight # (K, D)
    l_reg = Reg(g1_w, g2_w)
    
    if pro < T:
        L = l_bias + l_reg # stage 1
    else:
        L = l_bias + l_reg + l_gen # stage 2

    L.backward()  # back-propagate
    update(f, cls_g1, cls_g2)  # SGD update

def Reg(w1, w2):  # orthogonality regulation
    w1 = normalize(w1, dim=-1)  # l2-normalize
    w2 = normalize(w2, dim=-1)  # l2-normalize
    reg = einsum("kmd,kd->km", w1, w2)
    return mean(sum(reg ** 2, dim=-1))
\end{lstlisting}
\end{algorithm}

\section{Experiments}

\par We evaluate the effectiveness of MAD for single domain generalization (single-DG) via various empirical evidences on a series of tasks, including recognition on images, point clouds, texts, and semantic segmentation on images. Here we include several single-DG methods as baselines for performance comparison: (1) ERM~\cite{vapnik_erm} directly applies the vanilla strategy to train source model. (2) AugMix~\cite{hendrycks2020augmix} utilizes stochastic and diverse augmentations, and a formation to mix multiple augmented images to generate diverse samples. (3) pAdaIN~\cite{nuriel2021pAaIN} swaps feature statistics between the samples applied with a random permutation of mini-batch, (4) Mixstyle~\cite{mixstyle} adopts linear interpolation on feature statistics of two instances to generate synthesized samples. (5) DSU~\cite{DSU} characterizes the feature statistics as uncertain distribution to model domain shift. (6) ACVC~\cite{acvc} introduces more severe image augmentations, including image corruptions and Fourier transform. Recall that our MAD is able to directly strengthen the capability of classifier to identify domain-specific features, and meanwhile emphasize the learning of domain-generalized features. Therefore, MAD can be seamlessly incorporated into these methods to further boost performances. Note that MAD discards the additional biased-branch and only employs the feature extractor plus general-branch classifier at inference. That is, when plugging MAD into existing methods, there is no increase in computational cost.


\subsection{Single-DG on Image Recognition}
\noindent \textbf{Setup and Implementation Details:} We validate the proposed method on two image datasets: \textbf{PACS}~\cite{li2017pacs}, a widely-used benchmark for domain generalization with four domains: Photo (P), Art Painting (A), Cartoon (C), and Sketch (S). \textbf{VLCS}~\cite{torralba2011vlcs}, another commonly adopted benchmark for domain generalization with four different domains: VOC2007 (V), LabelMe (L), Caltech101 (C), SUN09 (S). In our implementation, we adopt the ResNet-18~\cite{resnet} pretrained on ImageNet~\cite{imagenet} as backbone. We apply the SGD optimizer with momentum 0.9. The batch size is set to 64. We set the learning rate to 2e-3/1e-3 for PACS/VLCS. Experiments are conducted on a Tesla P40 GPU with PyTorch-1.5. 
Following~\cite{domain_bed}, we split the training domain into training and validation subsets, and select the best-performing model on validation set to report the OOD performances.
\\
\noindent \textbf{Experiment Results:} We first conduct experiments on PACS, shown in Table~\ref{tab:pacs}. The main domain shift in this dataset is derived from style differences, 
and most data augmentation methods manifest higher performances than ERM baseline. Though these methods have achieved good performances, our MAD still manages to further improve their performance consistently. For example, MAD boosts up the overall accuracy of ACVC from 63.61\% to 65.87\%. Table~\ref{tab:vlcs} further summarizes the performance comparison on VLCS. The domain shift of this dataset mainly comes from background and view point changes. The scenes in VLCS vary from urban to rural, and the viewpoint tends to favor the side view or non-classical view. As a result, existing data augmentation methods which mainly introduce diverse styles obtain relatively smaller performance gains on VLCS dataset compared to those on PACS. Even in this case, MAD significantly improves the overall accuracy of ERM from 59.56\% to 62.95\%. Especially when taking ``LabelMe" as source domain, our MAD leads to near 10\% improvement in average accuracy. Similar to the observations on PACS, the consistent performance improvements are attained when integrating existing data augmentation approaches with MAD. In particular, MAD increases the accuracy of ACVC from 61.25\% to 63.82\%. 

\begin{table}[tbp]
  \caption{Single-domain generalization classification accuracies (\%) on PACS dataset with ResNet-18 as backbone. Here, P, A, C, and S denote the source domains. We train the model on one source domain, and evaluate them on the rest domains.}
  \vspace{-0.05in}
  \addtolength{\tabcolsep}{-2.5pt}
  \resizebox{0.47\textwidth}{!}{
    \begin{tabular}{lccccca}
    \toprule
    Methods & {Venue} & P     & A     & C     & S     & Avg \\
    \midrule
    ERM   &       & 33.65  & 65.38  & 64.20  & 34.15  & 49.34  \\
    ERM w/ MAD &       & 32.32  & 66.47  & 69.80  & 34.54  & \textbf{50.78} \\
    \midrule
    Augmix~\cite{hendrycks2020augmix} & {\multirow{2}[0]{*}{ICLR 19}} & 38.30  & 66.54  & 70.16  & 52.48  & 56.87  \\
    Augmixw/ MAD &       & 36.19  & 68.04  & 73.11  & 54.44  & \textbf{57.94} \\
    \midrule
    pAdaIn~\cite{nuriel2021pAaIN} & {\multirow{2}[0]{*}{CVPR 21}} & 33.66  & 64.96  & 65.24  & 32.04  & 48.98  \\
    pAdaIn w/ MAD &       & 34.66  & 65.64  & 70.10  & 42.85  & \textbf{53.31} \\
    \midrule
    Mixstyle~\cite{mixstyle} & {\multirow{2}[0]{*}{ICLR 21}} & 37.44  & 67.60  & 70.38  & 34.57  & 52.50  \\
    Mixstyle w/ MAD &       & 41.57  & 69.88  & 71.61  & 41.58  & \textbf{56.16} \\
    \midrule
    ACVC~\cite{acvc}  & {\multirow{2}[0]{*}{CVPR 22}} & 48.05  & 73.68  & 77.39  & 55.30  & 63.61  \\
    ACVC w/ MAD &       & 52.95  & 75.51  & 77.25  & 57.75  & \textbf{65.87} \\
    \midrule
    DSU~\cite{DSU}   & {\multirow{2}[0]{*}{ICLR 22}} & 42.10  & 71.54  & 74.51  & 47.75  & 58.97  \\
    DSU w/ MAD &       & 44.15  & 72.41  & 74.47  & 49.60  & \textbf{60.16} \\
    \bottomrule
    \end{tabular}%
    }
  \label{tab:pacs}%
\end{table}%

\begin{table}[tbp]
  \centering
  \caption{Single-domain generalization classification accuracies (\%) on VLCS dataset with ResNet-18 as backbone. Here, V, L, C, and S denote the source domains. We train the model on one source domain, and evaluate them on the rest domains.}
  \vspace{-0.05in}
  \addtolength{\tabcolsep}{-2.5pt}
  \resizebox{0.47\textwidth}{!}{
    \begin{tabular}{lccccca}
    \toprule
    Methods & Venue & V     & L     & C     & S     & Avg \\
    \midrule
    ERM   &       & 76.72  & 58.86  & 44.95  & 57.71  & 59.56  \\
    ERM w/ MAD &       & 76.21  & 67.97  & 46.55  & 61.04  & \textbf{62.95} \\
    \midrule
    Augmix~\cite{hendrycks2020augmix} & \multirow{2}[2]{*}{ICLR 19} & 75.25  & 59.52  & 45.90  & 57.43  & 59.53  \\
    Augmixw/ MAD &       & 76.57  & 65.60  & 44.35  & 59.47  & \textbf{61.50} \\
    \midrule
    pAdaIn~\cite{nuriel2021pAaIN} & \multirow{2}[2]{*}{CVPR 21} & 76.03  & 65.21  & 43.17  & 57.94  & 60.59  \\
    pAdaIn w/ MAD &       & 76.57  & 68.90  & 42.92  & 63.91  & \textbf{63.08} \\
    \midrule
    Mixstyle~\cite{mixstyle} & \multirow{2}[2]{*}{ICLR 21} & 75.73  & 61.29  & 44.66  & 56.57  & 59.56  \\
    Mixstyle w/ MAD &       & 75.00  & 66.17  & 43.61  & 62.01  & \textbf{61.70} \\
    \midrule
    ACVC~\cite{acvc}  & \multirow{2}[2]{*}{CVPR 22} & 76.15  & 61.23  & 47.43  & 60.18  & 61.25  \\
    ACVC w/ MAD &       & 76.15  & 69.36  & 48.04  & 61.74  & \textbf{63.82} \\
    \midrule
    DSU~\cite{DSU}   & \multirow{2}[2]{*}{ICLR 22} & 76.93  & 69.20  & 46.54  & 58.36  & 62.76  \\
    DSU w/ MAD &       & 76.99  & 70.85  & 44.78  & 62.23  & \textbf{63.71} \\
    \bottomrule
    \end{tabular}%
    }
  \label{tab:vlcs}%
  \vspace{-0.20in}
\end{table}%

\subsection{Single-DG on Point Cloud Recognition}
\noindent \textbf{Setup and Implementation Details:} Different from 2D vision, 3D vision has various modalities to represent data, such as voxel grid, 3D mesh and point cloud. Among them, point cloud is the most straightforward and representative modality, which consists of a set of points with 3D coordinates. To verify the generality of MAD, we conduct experiments on the 3D point cloud domain adaptation dataset \textbf{PointDA-10}~\cite{pointdan}, which consists of three domains: ShapeNet (SH), ScanNet (SC), and ModelNet (M). In our implementation, we adopt the PointNet~\cite{pointdan} as backbone, and apply the SGD optimizer with momentum 0.9. The batch size is set to 64. We set the learning rate to 1e-3. Experiments are executed on a Tesla P40 GPU with PyTorch-1.5. For model selection, similar to the experiments on images, we split the training domain into training and validation subsets, and choose the model with maximal accuracy on validation set to report the OOD performance.

\noindent \textbf{Experiment Results:} 
Table~\ref{tab:pointda} lists the performance comparison on PointDA-10. An observation is that the existing data augmentation methods on 2D images do not work well on 3D point clouds. The representative methods, e.g., Mixstyle and DSU, are even inferior to ERM. We speculate that this may be the results of the different types of domain shifts, which typically lie in geometric differences in point clouds rather than texture and style differences in 2D images. Moreover, there is no one-to-one correspondence and order between points, making it difficult to directly generate new point clouds by interpolating two point clouds. This somewhat reveals the weakness of data augmentation, when generalizing to different modalities. MAD, in comparison, benefits from decoupling domain-specific features and domain-generalized features, and constantly enhances these methods. In particular, MAD improves the overall accuracy of ERM/Mixstyle/DSU from 34.57\%/33.78\%/33.63\% to 37.91\%/38.16\%/36.45\%. The results basically indicate the advantage of MAD across different modalities.

\begin{table}[tbp]
  \centering
  \vspace{-0.05in}
  \caption{Single-domain generalization classification accuracies (\%) on PointDA-10 dataset with PointNet as backbone. Here, SH, SC, and M denote the source domains. We train the model on one source domain, and evaluate them on the rest domains.}
  \addtolength{\tabcolsep}{-2.0pt}
  \resizebox{0.47\textwidth}{!}{
    \begin{tabular}{lcccca}
    \toprule
    Methods & Venue & SH    & SC    & M     & Avg \\
    \midrule
    ERM   & \multirow{2}[0]{*}{-} & 25.69  & 45.09  & 32.94  & 34.57  \\
    ERM w/ MAD &       & 31.11  & 48.07  & 34.69  & \textbf{37.91}  \\
    \midrule
    Mixstyle~\cite{mixstyle} & \multirow{2}[0]{*}{ICLR 21} & 27.18  & 46.25  & 27.93  & 33.78  \\
    Mixstyle w/ MAD &       & 29.89  & 51.01  & 33.57  & \textbf{38.16}  \\
    \midrule
    DSU~\cite{DSU}   & \multirow{2}[1]{*}{ICLR 22} & 25.74  & 43.53  & 31.61  & 33.63  \\
    DSU w/ MAD &       & 28.92  & 47.69  & 32.72  & \textbf{36.45}  \\
    \bottomrule
    \end{tabular}%
    }
 \vspace{-0.15in}
 \label{tab:pointda}%
\end{table}%

\subsection{Single-DG on Text Classification}
\noindent \textbf{Setup and Implementation Details:} In addition to 2D images, and 3D point clouds, we further conduct experiments on cross-domain text classification. We choose the \textbf{Amazon Reviews}~\cite{amazon_review} as the benchmark, which contains four different domains on product review, including DVDs (D), Kitchen appliance (K), Electronics (E), and Books (B). The dataset has already been pre-processed into a bag of features (unigrams and bigrams), losing all word order information. Following~\cite{nlp_mdtc, nlp_cltl}, we take the 5,000 most frequent features and represent each review as a 5,000-dimentional feature vector. 
Following~\cite{nlp_mdtc, nlp_cltl}, we employ an MLP as feature extractor. We apply the SGD optimizer with momentum 0.9. The batch size is set to 64. We set the learning rate to 1e-3. Experiments are conducted on a Tesla P40 GPU with PyTorch-1.5. We adopt the same model selection strategy as in image and point cloud recognition.
\\
\noindent \textbf{Experiment Results:} The results shown in Table~\ref{tab:text} clearly verify the effectiveness of MAD in comparison to the existing methods. Similar to the observations on 2D images and 3D point clouds, MAD also exhibits performance improvement to existing approaches on text modality. For example, MAD boosts up the accuracy of ERM on Books domain by 3.09\%, and leads to 0.94\%, 0.89\%, and 1.40\% gain in overall accuracy to Mixup, Mixstyle, and DSU, respectively. The improvements empirically prove the impact of MAD on text modality.

\begin{table}[tbp]
  \centering
  \caption{Single-domain generalization classification accuracies (\%) on Amazon-Review dataset with an MLP as backbone. Here,  D, E, K, and B denote the source domains. We train the model on one source domain, and evaluate them on the rest domains.}
  \vspace{-0.10in}
  \addtolength{\tabcolsep}{-3.5pt}
  \resizebox{0.47\textwidth}{!}{
    \begin{tabular}{lccccca}
    \toprule
    Methods & {Venue} & {D} & {E} & {K} & {B} & {Avg} \\
    \midrule
    ERM   &  \multirow{2}[0]{*}{-}  & 74.17  & 73.17  & 73.67  & 71.58  & 73.15  \\
    ERM w/ MAD &       & 76.08  & 74.33  & 73.33  & 74.67  & \textbf{74.60} \\
    \midrule
    Mixup~\cite{zhang2017mixup} & \multirow{2}[0]{*}{ICLR 18} & 74.83  & 72.17  & 73.58  & 72.67  & 73.31  \\
    Mixup w/ MAD &       & 75.33  & 73.58  & 74.33  & 73.75  & \textbf{74.25} \\
    \midrule
    Mixstyle~\cite{mixstyle} & \multirow{2}[0]{*}{ICLR 21} & 74.75  & 73.17  & 74.33  & 72.33  & 73.65  \\
    Mixstyle w/ MAD &       & 75.17  & 72.75  & 75.00  & 75.25  & \textbf{74.54} \\
    \midrule
    DSU~\cite{DSU}   & \multirow{2}[1]{*}{ICLR 22} & 75.00  & 73.45  & 75.25  & 73.08  & 74.20  \\
    DSU w/ MAD &       & 76.42  & 74.33  & 76.50  & 75.17  & \textbf{75.60} \\
    \bottomrule
    \end{tabular}%
    }
    \vspace{-0.10in}
  \label{tab:text}%
\end{table}%

\begin{table}[tbp]
  \centering
  \caption{Single-domain generalization on semantic segmentation (GTA-5 $\rightarrow$ Cityscapes).}
  \vspace{-0.10in}
  \addtolength{\tabcolsep}{4.0pt}
  \resizebox{0.47\textwidth}{!}{
    \begin{tabular}{lccc}
    \toprule
    Methods & {Venue} & {mIOU(\%)} & {mACC(\%)} \\
    \midrule
    ERM   &  -     & 37.0  & 51.5  \\
    pAdaIN~\cite{nuriel2021pAaIN} & CVPR 21 & 38.3  & 52.1  \\
    Mixstyle~\cite{mixstyle} & ICLR 21 & 40.3  & 53.8  \\
    DSU~\cite{DSU}   & ICLR 22 & 42.3  & 54.7  \\
    \midrule
    ERM w/ MAD &  -     & \textbf{38.9} & \textbf{52.2}  \\
    DSU w/ MAD &  -    & \textbf{43.8} & \textbf{57.2} \\
    \bottomrule
    \end{tabular}%
    }
  \label{tab:segmentation}%
  \vspace{-0.20in}
\end{table}%
\subsection{Single-DG on Semantic Segmentation}
\noindent \textbf{Setup and Implementation Details:} The aforementioned experiments mainly focus on the single-DG recognition of 1D texts, 2D images and 3D point clouds. In this section, we experiment with 2D images segmentation. As a fundamental ability for autonomous driving, semantic segmentation models often encounter severe performance degeneration due to scenarios change. Here, we conduct experiments on GTA-5~\cite{gta5_dataset} $\rightarrow$ Cityscape~\cite{cordts2016cityscapes} datasets, the most widely-used benchmark on semantic segmentation domain adaptation. The experiments are based on FADA released codes~\cite{FADA}, using DeepLab-V2~\cite{chen2017deeplab} segmentation network with ResNet-101~\cite{resnet} as backbone. We apply the SGD optimizer with momentum 0.9. The batch size is set to 8. We set the learning rate to 5e-4. Experiments are implemented on 4 Tesla P40 GPUs with PyTorch-1.5. Mean Intersection over Union (mIOU) and mean Accuracy (mAcc) for all objects categories are adopted as evaluation metric. 

\noindent \textbf{Experiment Results:} As a pixel-level classification task, semantic segmentation is much harder than image-level recognition. Table~\ref{tab:segmentation} details the results, demonstrating the superiority of MAD against baselines. Specifically, MAD contributes an mIOU increase of 1.9\% and 1.5\% to ERM and DSU, respectively. The results again verify the merit of MAD on semantic segmentation on 2D images.

\subsection{Experiments Analysis}
\label{sec:exp_ablation_hyper_param}

\begin{table}[tbp]
 \centering
  \caption{\textbf{Ablation}. Results of the vanilla ERM, ERM w/ MAD (one-stage), ERM w/ MAD (single-head), and ERM w/ MAD. The experiments are conducted on single-domain generalization scenarios of 1D Texts (Amazon Review dataset), 2D Images (VLCS dataset), and 3D Point Clouds (PointDA-10 dataset).}
  \addtolength{\tabcolsep}{-5.0pt}
  \resizebox{0.47\textwidth}{!}{
    \begin{tabular}{lcccc}
    \toprule
    Methods &{1D Texts} & {2D Images} & {3D Points} & Avg \\
    \midrule
    ERM   & 73.15  & 59.56  & 34.57  & 55.76  \\
    MAD (one-stage) & 74.00  & 60.41  & 35.67  & 56.69  \\
    MAD (single-head) & 74.31  & 60.49  & 36.61  & 57.14  \\
    \midrule
    MAD & \textbf{74.60}  & \textbf{62.95}  & \textbf{37.91}  & \textbf{58.49}  \\
    \bottomrule
    \end{tabular}%
    }
  \label{tab:ablation}%
  \vspace{-0.10in}
\end{table}%
\noindent \textbf{Ablation Study:} To examine the contribution of different components within MAD, we first conduct extensive ablation studies on texts, images, and point clouds recognition. Table~\ref{tab:ablation} summarizes the results. Here, \textit{MAD (one-stage)} refers to a degraded version of MAD without two-stage learning mechanism. That is, we optimize the biased-branch $g_{bias}$ and the general-branch $g_{gen}$ simultaneously. \textit{MAD (single-head)} indicates that we only capitalize on a single-head classifier in the biased-branch to capture those domain-specific features. As shown in Table~\ref{tab:ablation}, both the multi-head classifier design and the two-stage learning mechanism are effective. The two components complement to each other and both manage the general-branch classifier $g_{gen}$ to focus more on those domain-generalized features.
\begin{figure}[t]
    \centering
    \includegraphics[width=0.45\textwidth]{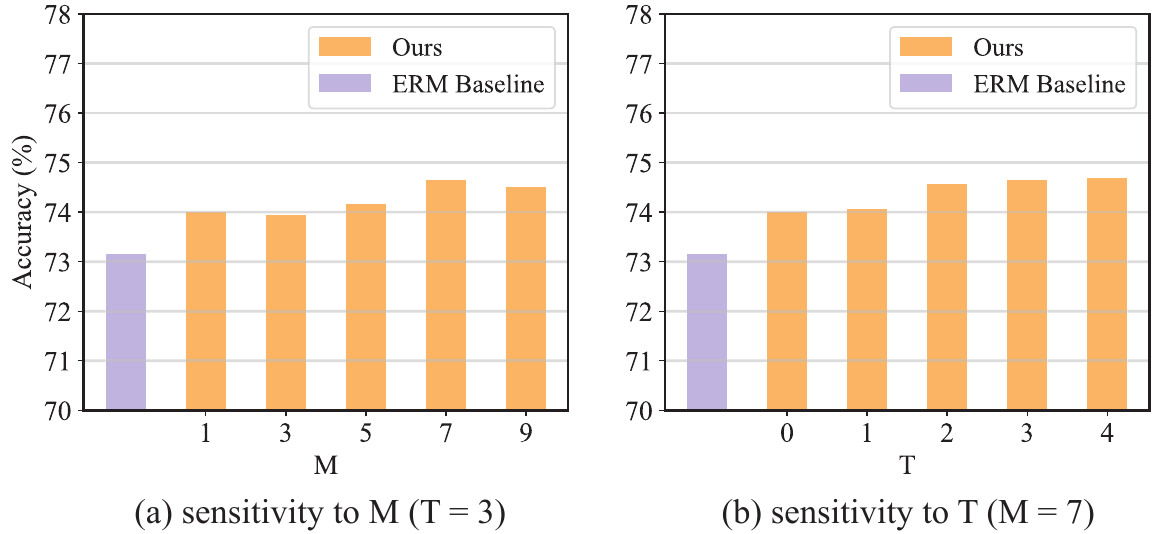}
    \vspace{-0.05in}
    \caption{Hyper-parameter sensitivity analysis on Single-DG text classification.  M denotes the number of the biased-branch classifiers and T is the training threshold for the second stage.}
    \vspace{-0.20in}
    \label{fig:hyper_param}
\end{figure}
\\
\noindent \textbf{Hyper-parameter Sensitivity:} Next, we study the hyper-parameter sensitivity of $M$ and $T$ on text classification task. $M$ is the number of the biased-branch classifiers, and $T$ denotes the second-stage training threshold. 
As shown in Figure~\ref{fig:hyper_param}, the accuracies are relatively stable when each hyper-parameter varies. In our implementation, we set $T$ to 3. Since $M$ depends on the factors of the introduced domain-specific features, its value differs for different datasets. We set $M$ to 3 for PointDA-10 and GTA-5, 5 for VLCS, and 7 for PACS and Amazon Review.

\begin{figure}
    \centering
    \includegraphics[width=0.47\textwidth]{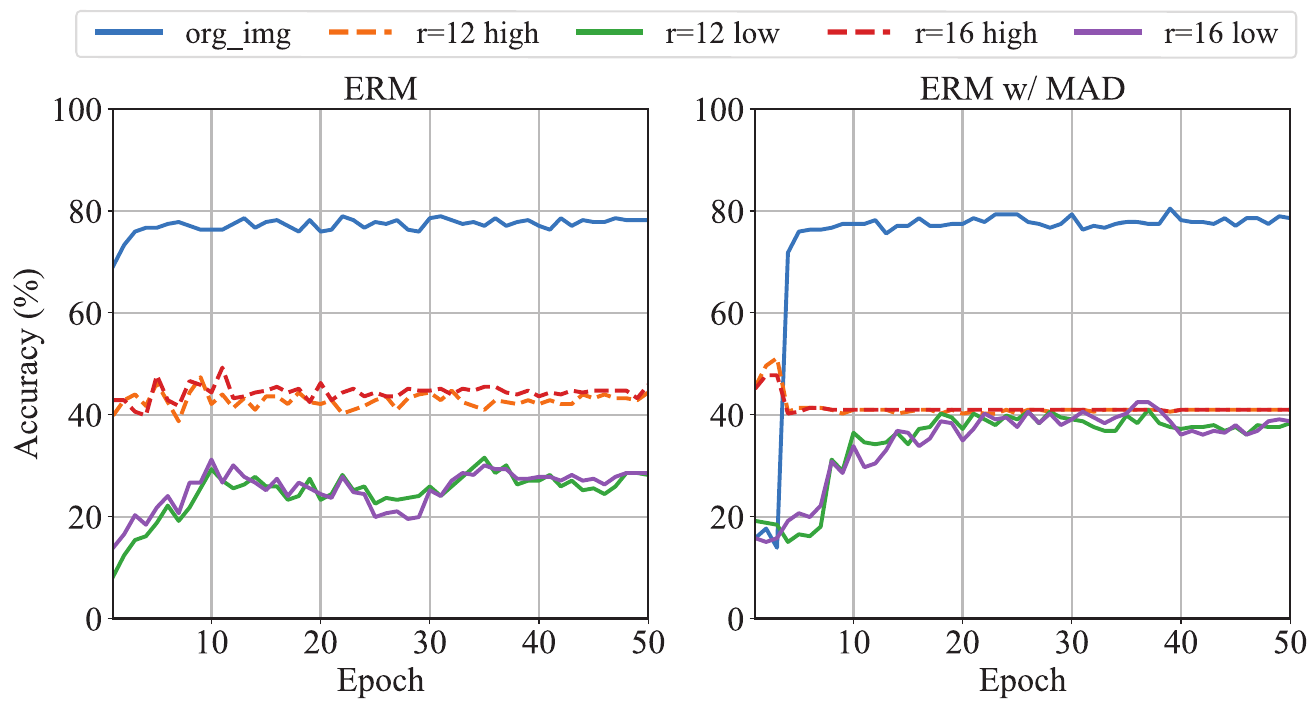}
    \vspace{-0.10in}
    \caption{Comparisons of ERM and ERM w/ MAD training curves on low-frequency component (LFC) and high-frequency component (HFC). Experiments are conducted in the ``LabelMe" domain of the VLCS benchmark. All curves in this figure are from validation samples of ``LabelMe" domain.}
    \vspace{-0.280in}
    \label{fig:lfc_hfc}
\end{figure}

\noindent \textbf{Low-frequency Component vs High-frequency Component:} As pointed out in~\cite{high_freq}, the low-frequency component (LFC) is much more generalizable than high-frequency component (HFC), i.e., LFC typically represents those domain-generalized (semantic) features, and HFC denotes those domain-specific (superficial) features. Here, we conduct experiments in the ``LabelMe" domain of the VLCS benchmark to verify whether MAD encourages classifier to pay more attention to those domain-generalized features, i.e., the LFC. Specifically, for each instance in the validation subset, we decompose the data into LFC and HFC w.r.t different radius thresholds $r$ via applying Fourier transform and inverse Fourier transform. Then, we train the vanilla ERM classifier, the ERM classifier equipped with MAD separately, and evaluate them on LFC and HFC. Figure~\ref{fig:lfc_hfc} depicts the results, where $r=12/16\ low$ (solid line) denotes the LFC and $r=12/16\ high$ (dashed line) denotes the HFC. As shown in this figure, ERM w/ MAD performs much better on LFC than vanilla ERM, with an accuracy improvement of nearly 10\%. The results confirm the effectiveness of MAD in improving single-domain generalization, and MAD indeed encourages classifiers to pay more attention to those domain-generalized features (LFC).

\subsection{Visualization}
\begin{figure}
    \centering
    \includegraphics[width=0.47\textwidth]{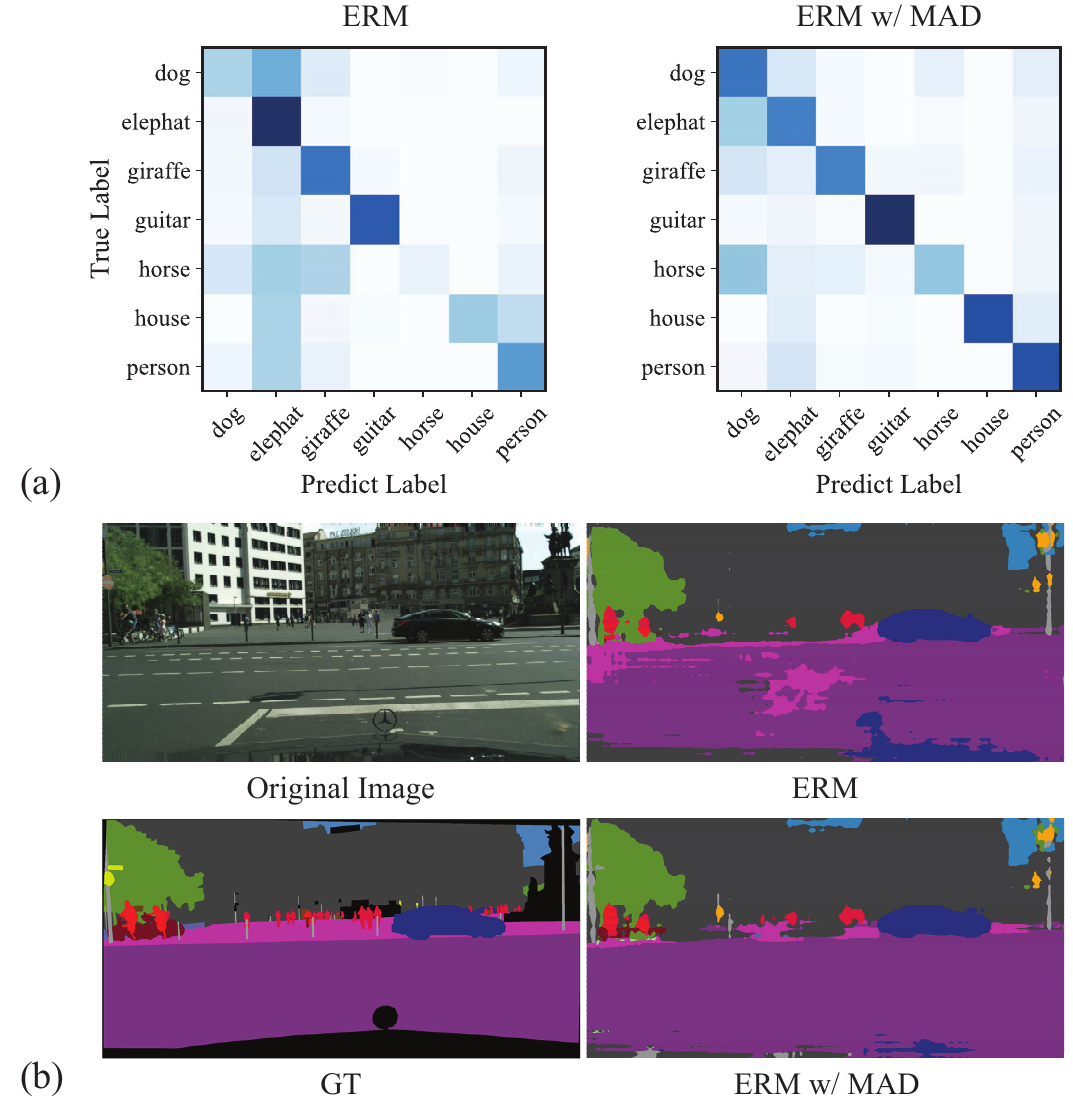}
    \vspace{-0.10in}
    \caption{\textbf{Visualization.} (a) Confusion matrix on PACS benchmark (``Carton"$\rightarrow$ ``Sketch"). (b) Semantic segmentation illustration on unseen domain Cityscapes with model trained on GTA-5.}
    \vspace{-0.350in}
    \label{fig:visuliazation}
\end{figure}

\par In addition to quantitative performance comparisons, we further present some qualitative illustrative results. Figure~\ref{fig:visuliazation} (a) first visualizes the confusion matrix on PACS benchmark. The classification model is trained on ``Cartoon" domain and evaluated on unseen domain ``Sketch". The results show that ERM w/ MAD is less confusing for most categories when testing in unseen domains compared to vanilla ERM. Then, Figure~\ref{fig:visuliazation} (b) illustrates an example for semantic segmentation. The visualization demonstrates that MAD can enhance the ERM baseline to achieve more precise segmentation results under domain shift, especially for the driveable areas.

\section{Conclusion}
\par In this paper, we delve into the single domain generalization (single-DG) problem. Different from existing methods that introduce modality-specific data augmentation techniques, we propose a general and versatile modality-agnostic debiasing (MAD) framework for single-DG. MAD starts from the viewpoint of directly strengthening the capability of classifier for identifying domain-specific (superficial) features, and meanwhile emphasizing the learning of domain-generalized (semantic) features. Technically, we have devised a novel two-branch classifier, where a biased-branch is responsible for identifying those superficial features, while the general-branch is encouraged to focus more on those semantic features. MAD is appealing in view that it can be seamlessly incorporated into existing methods to further boost up performances. We have evaluated the effectiveness and superiority of MAD for single-DG via various empirical evidences on a series of tasks, including recognition on 1D texts, 2D images, 3D point clouds, and semantic segmentation on 2D images. In all tasks, MAD can facilitate the state-of-the-art methods to achieve better performance without bells and whistles. 

\noindent\textbf{Acknowledgment:} 
This work was partially supported by Shanghai Municipal Science and Technology Major Project (No.2018SHZDZX01), ZJ Lab, and Shanghai Center for Brain Science and Brain-Inspired Technology, and the Shanghai Rising Star Program (No.21QC1400900).

\appendix
\begin{figure*}
  \centering
  \includegraphics[width=0.99\textwidth]{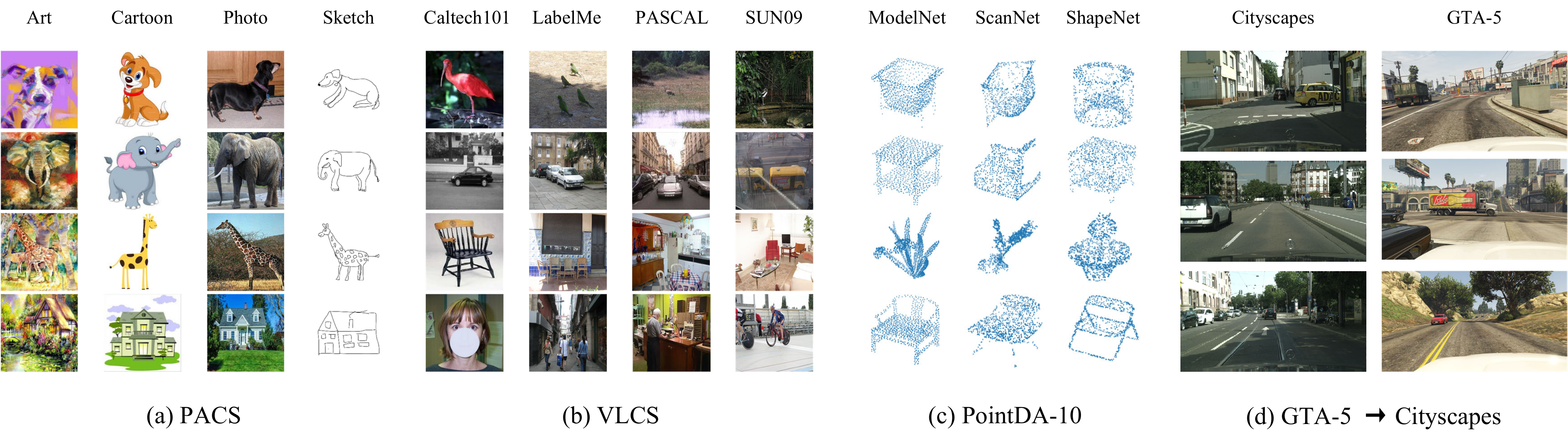}
   \captionof{figure}{Examples from four domain generalization benchmarks that manifest different types of domain shifts. In (a), image styles differences are the main source for domain shifts. In (b), the domain shifts mainly correspond to the changes of environments and viewpoints. In (c), the domain shifts are solely derived from the geometric differences. In (d), driving scenes changes are the main reason for domain shifts.}
  \vspace{-0.15in} 
\label{fig:dataset}
\end{figure*}

\begin{table*}[h!]
  \centering
  \caption{The statistics of benchmark datasets.}
  \vspace{-0.1in}
  \addtolength{\tabcolsep}{2.5pt}
  \resizebox{0.99\textwidth}{!}{
    \begin{tabular}{lcccp{7.5cm}l}
    \toprule
    Dataset & \multicolumn{1}{l}{\#Domain} & \multicolumn{1}{l}{\#Class} & \multicolumn{1}{l}{\#Sample} & Description & Reference\\
    \midrule
    PACS  & 4     & 7     & 9,991 & Art, Cartoon, Photos, Sketches. &\cite{li2017pacs}\\
    VLCS  & 4     & 5     & 10,729 & Caltech101, LabelMe, SUN09, VOC2007. &\cite{torralba2011vlcs}\\
    PointDA & 3     & 10    & 32,788 & ShapeNet, ScanNet, ModelNet. &\cite{pointdan}\\
    AmazonReview & 4     & 2     & 8,000 & DVDs, Kitchen, Electronics, Books. &\cite{amazon_review}\\
    GTA5$\rightarrow$ Cityscapes & 2     & 19    & 29,966 & Semantic segmentation generalization from synthetic images to realistic images. &\cite{gta5_dataset, cordts2016cityscapes}\\
    \bottomrule
    \end{tabular}%
    }    
  \label{tab:statistics}%
\end{table*}%

\begin{figure*}[h!]
    \centering
    \includegraphics[width=0.95\textwidth]{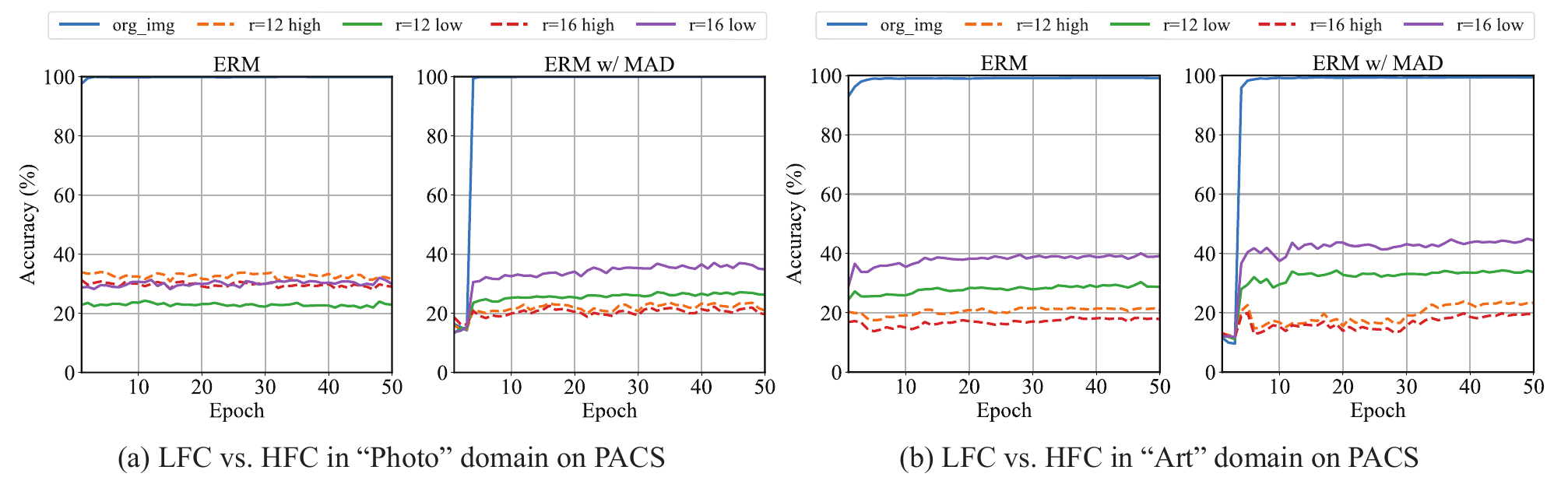}
    \vspace{-0.1in}
    \caption{Comparisons of ERM and ERM w/ MAD training curves on low-frequency component (LFC) and high-frequency component (HFC). Experiments are conducted on PACS. All curves in this figure are from validation samples.}
    \vspace{-0.1in}
    \label{fig:lfc_hfc_supp}
\end{figure*}

\begin{figure*}[h!]
    \centering
    \includegraphics[width=0.95\textwidth]{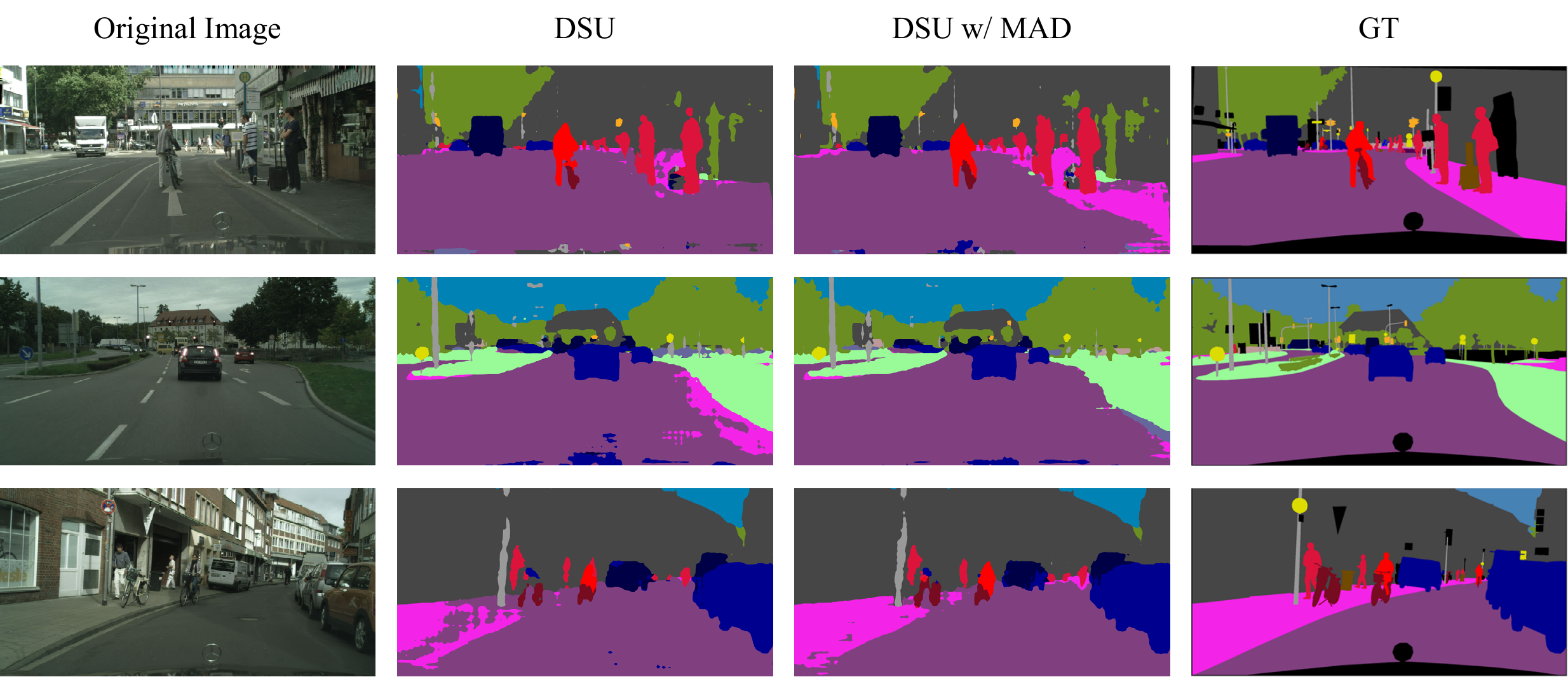}
    \caption{Semantic segmentation illustration on unseen domain Cityscapes with model trained on GTA-5.}
    \vspace{-0.1in}
    \label{fig:seg_supp}
\end{figure*}

\section{More Details about Datasets}
\par In the main paper, we have validated the effectiveness of our Modality-Agnostic Debiasing (MAD) framework in a variety of single domain generalization (single-DG) scenarios with different modalities, including recognition on 1D texts, 2D images, 3D point clouds, and semantic segmentation on 2D images. Here we provide more details about the adopted datasets in the main paper. The statistics are listed in Table~\ref{tab:statistics}.

\par In an effort to qualitatively show the domain shifts in different benchmarks, we further illustrate some examples in Figure~\ref{fig:dataset}. One major observation is that the domain shifts vary a lot between benchmarks. For example, the domain shifts in images (Figure~\ref{fig:dataset} (a), (b), (d)) mostly result from the changes for image contexts, styles, and viewpoints. In point clouds (Figure~\ref{fig:dataset} (c)), the domain shifts primary correspond to geometric variations. Existing single-DG methods are commonly designed for images by devising various data augmentation algorithms to introduce various textures and image styles, making them modality-specific and only applicable to the single modality inputs of images. In contrast, MAD proposes to directly enhance the classifier's ability to identify domain-specific features while emphasizing the learning of domain-generalized features. In this way, a versatile modality-agnostic single-DG paradigm is established by completely eliminating the need for modality-specific data augmentations. MAD is also appealing due to the fact that it can be seamlessly incorporated into existing single-DG methods to further boost up performances.

\section{More Results for Low-Frequency Component vs. High-Frequency Component}
\par For images, Low-frequency component (LFC) is commonly considered as domain-generalized features, while High-frequency component (HFC) is regarded as domain-specific features~\cite{high_freq}. Here, we provide more results to support the capacity of MAD enforcing classifiers to pay more attention to domain-generalized features, i.e., LFC. Here we conduct additional experiments in the ``Photo" and ``Art" domains on PACS benchmark. Implementation details are the same as in the main paper. That is, for each instance in the validation subset, we decompose the image into LFC and HFC w.r.t different radius threshold $r$ via applying Fourier transform and inverse Fourier transform. Then, we train the ERM and the ERM w/ MAD, separately, and evaluate them on LFC and HFC. The results are summarized in Figure~\ref{fig:lfc_hfc_supp}, where $r= 12/16 low$ represents the LFC and $r=12/16 high$ depicts the HFC. As shown in this figure, we can conclude that MAD consistently encourages the classifier focus more on those domain-generalized features.

\section{More Results for Semantic Segmentation Visualization}
\par Semantic segmentation models often suffer from performance degradation due to scenario changes.  We exhibit more visualization results of semantic segmentation in Figure ~\ref{fig:seg_supp}. These examples further demonstrate the effectiveness of MAD when integrated into existing data-augmentation based methods (e.g., DSU~\cite{DSU}).

{\small
\bibliographystyle{ieee_fullname}
\bibliography{egbib}
}

\end{document}